\documentclass{article} 
\usepackage{iclr2018_conference,times}
\usepackage{hyperref}
\usepackage{url}
\usepackage[utf8]{inputenc} 
\usepackage[T1]{fontenc}    
\usepackage{hyperref}       
\hypersetup{ hidelinks, }
\usepackage{longtable}      
\usepackage{booktabs}       
\usepackage{amsfonts}       
\usepackage{nicefrac}       
\usepackage{microtype}      
\usepackage[pdftex]{graphicx}
\usepackage{xcolor}
\usepackage{array,multirow,graphicx}
\newcommand\bigcdot{\boldsymbol{{}\cdot{}}}
\newcommand*\rot{\rotatebox{67}}
\usepackage{amsmath,amssymb}
\usepackage[colorinlistoftodos,prependcaption,textsize=tiny]{todonotes}

\title{Ask the Right Questions:\\Active Question Reformulation with\\Reinforcement Learning}

\author{Christian Buck, Jannis Bulian, Massimiliano Ciaramita, Wojciech Gajewski,\\ {\bf Andrea Gesmundo, Neil Houlsby, Wei Wang}\\
Google\\
\footnotesize{\texttt{\{cbuck,jbulian,massi,wgaj,agesmundo,neilhoulsby,wangwe\}@google.com}} \\
}

\iclrfinalcopy

\newcommand{\ra}[1]{\renewcommand{\arraystretch}{#1}}
\begin{document}

\maketitle

\begin{abstract}
  We frame Question Answering (QA) as a Reinforcement Learning task, an
approach that 
we call \emph{Active Question Answering}. We propose an agent
that sits between 
the user and a black box QA system and learns to
reformulate questions to elicit the best possible answers. The agent probes the
system with, potentially many, natural language reformulations of an initial
question and aggregates the returned evidence to yield the best answer.
The reformulation system is trained end-to-end to maximize answer quality using
policy gradient.
We evaluate on SearchQA, a dataset of complex questions
extracted from \emph{Jeopardy!}.
The agent outperforms a state-of-the-art base model, playing the
role of the environment, and other benchmarks.
We also analyze the language that the agent has learned while
interacting with the question answering system.
We find that successful question reformulations look quite
different from natural language paraphrases.
The agent is
able to discover non-trivial reformulation strategies that resemble
classic information retrieval techniques such as term re-weighting
(tf-idf) and stemming. 
 \end{abstract}
\section{Introduction}

Web and social media have become primary sources of information.
Users' expectations and information seeking activities co-evolve
with the increasing sophistication of these resources.
Beyond navigation, document retrieval, and simple factual question answering,
users seek direct answers to complex and compositional questions. Such
search sessions may require multiple iterations, critical assessment, and
synthesis~\citep{Marchionini:2006}.

The productivity of natural language yields a myriad of ways to formulate a
question~\citep{Chomsky:1965}.
In the face of complex information needs,
humans overcome uncertainty by reformulating
questions, issuing multiple searches, and aggregating responses.
Inspired by humans' ability to \emph{ask the right questions},
we present an agent that learns to carry out this process for the user.
The agent sits between the user and a backend QA system
that we refer to as `the environment'. We call the agent AQA, as it
implements an \emph{active question answering} strategy.
AQA aims to maximize the chance of getting the correct answer by
sending a reformulated question to the environment.
The agent seeks to find the best answer by asking many
questions and aggregating the returned evidence.
The internals of the environment are not available to the agent, so it
must learn to probe a black-box optimally using only question strings.
The key component of the AQA agent is a
sequence-to-sequence model
trained with reinforcement learning (RL) using a reward based on the answer
returned by the environment. The second component to AQA combines the
evidence from interacting with the environment using a convolutional
neural network to select an answer.

We evaluate on a dataset of
\emph{Jeopardy!} questions, SearchQA~\citep{Dunn:2017}.
These questions are hard to answer by design because they use
convoluted language,
e.g., \emph{Travel doesn't seem to be an issue for this sorcerer
\& onetime surgeon; astral projection \& teleportation are no prob}
(answer: \emph{Doctor Strange}).
Thus SearchQA tests the ability of AQA to reformulate questions such that the
QA system has the best chance of returning the correct answer.
AQA improves over the performance of a deep network built for QA,
BiDAF~\citep{Seo:2017}, which has produced state-of-the-art results on
multiple tasks, by 11.4\% absolute F1,
a 32\% relative F1 improvement. Additionally, AQA outperforms other competitive
heuristic query reformulation benchmarks.

AQA defines an instance of machine-machine communication.
One side of the conversation, the AQA agent, is trying to adapt its
language to improve the response from the other side, the QA
environment.
To shed some light on this process we perform a qualitative analysis
of the language  generated by the AQA agent. By evaluating on
MSCOCO~\citep{MSCOCO:2014}, we find that the agent's question
reformulations diverge significantly from natural language
paraphrases. Remarkably, though, the agent is able to learn non-trivial and
transparent policies. In particular, the agent is able to
discover classic IR query operations such as term re-weighting,
resembling tf-idf, and morphological simplification/stemming.
A possible reason being that
current machine comprehension tasks involve the ranking of short textual
snippets, thus incentivizing relevance, more than deep language understanding.
 \section{Related work}
\citet{Lin:2001} learned patterns of
question variants by comparing dependency parsing trees.
\citet{Duboue:2006} showed that MT-based paraphrases can be useful in
principle by providing significant headroom in oracle-based
estimations of QA performance. Recently, \citet{Berant:2014} used
paraphrasing to augment the training of a semantic parser by expanding
through the paraphrases as a latent representation.
Bilingual corpora and MT have been used to generate
paraphrases by pivoting through a second language.
Recent work
uses neural translation models and multiple
pivots~\citep{mallinsonparaphrasing}. 
In contrast, our approach does not use pivoting and is, to our knowledge, the
first direct neural paraphrasing system.
\citet{Riezler2007QueryExpansion} propose phrase-based paraphrasing
for query expansion.
In contrast with this line of work, our goal is to generate full
question reformulations while optimizing directly the end-to-end target performance
metrics.

Reinforcement learning
is gaining traction in natural language understanding across many problems.
For example,~\citet{Narasimhan:2015} use RL
to learn control policies for multi-user dungeon games
where the state of the game is summarized by a textual description,
and \citet{Li:2017} use RL for dialogue generation.
Policy gradient methods have been investigated recently for MT and
other sequence-to-sequence problems.
They alleviate limitations inherent to the word-level optimization of the
cross-entropy loss, allowing the use of sequence-level reward functions, like BLEU.
Reward functions based on language models and
reconstruction errors are used to bootstrap MT with fewer resources
\citep{Xia:2016}.
RL training can also prevent \emph{exposure bias};
an inconsistency between training and inference time
stemming from the fact that the model never sees its own
mistakes during training~\citep{Ranzato:2015}.
We also use policy gradient to optimize our agent, however, we
use end-to-end question answering quality as the reward.

Uses of policy gradient for QA include \citet{Liang:2017},
who train a semantic parser to query a knowledge base, and
\citet{Seo:2017b} who propose query reduction networks
that transform a query to answer questions that involve
multi-hop common sense reasoning.
The work of \citet{Nogueira:2016} is most related to ours.
They identify a document containing
an answer to a question by following links on a graph.
Evaluating on a set of questions from the game \emph{Jeopardy!},
they learn to walk the Wikipedia graph until
they reach the predicted answer.
In a follow-up, \citet{Nogueira:2017} improve document
retrieval with an approach inspired by relevance feedback
in combination with RL.
They reformulate a query by adding terms from documents retrieved
from a search engine for the original query.
Our work differs in that we generate
complete sequence reformulations rather than adding single
terms, and we target question-answering rather than document
retrieval.

Active QA is also related to recent research on
fact-checking:~\citet{Wu:2017} propose to perturb database queries in
order to estimate the support of quantitative claims. In Active QA
questions are perturbed semantically with a similar purpose, although
directly at the surface natural language form. 
 \section{Active Question Answering Model}\label{sec:AQAmodel}
\begin{figure}
\centering
\includegraphics[width=0.5\linewidth]{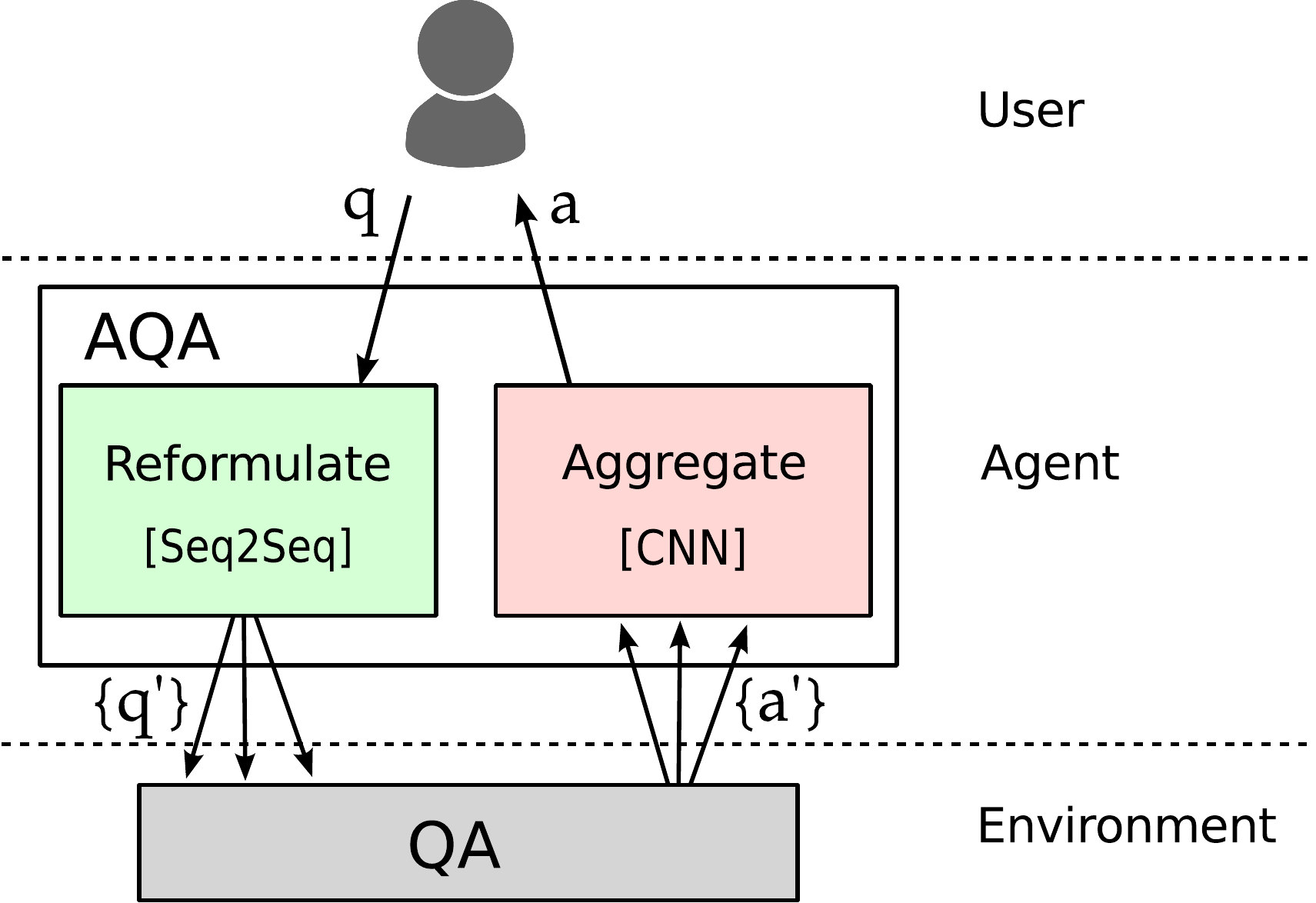}
\caption{The AQA agent-environment setup.
	 In the downward pass the agent reformulates the question and sends
	 variants to the QA system. In the upward pass the final answer
	 is selected.}
\label{fig:aqa}
\end{figure}

Figure~\ref{fig:aqa} shows the Active Question Answering (AQA)
agent-environment setup.
The AQA model interacts with a black-box environment. AQA queries it with
many versions of a question, and finally returns the best of the
answers found. 
An episode starts with an original question $q_0$. The agent then generates a
set of reformulations $\{q_i\}_{i=1}^N$.
These are sent to the environment which returns answers $\{a_i\}_{i=1}^N$.
The selection model then picks the best from these candidates.

\subsection{Question-Answering Environment} \label{sec:QAEnvironment}
For the QA environment, we use a competitive neural question answering model,
BiDirectional Attention Flow (BiDAF)~\citep{Seo:2017}.
BiDAF is an extractive QA system,
it selects answers from contiguous spans of a given document.
Given a question, the environment returns an answer and, during training,
a reward.
The reward may be any quality metric for the returned answer,
we use token-level F1 score.
Note that the reward for each answer $a_i$
is computed against the original question $q_0$.
We assume that the environment is opaque;
the agent has no access to its parameters, activations or gradients.
This setting enables one, in principle, to also interact with other
information sources, possibly providing feedback in different modes
such as images and structured data from knowledge bases.
However, without propagating gradients through the environment we lose
information, 
feedback on the quality of the question
reformulations is noisy, presenting a challenge for training.

\subsection{Reformulation Model}
The reformulator is a sequence-to-sequence model, as is popular for
neural machine translation.
We build upon the implementation of~\citet{Britz:2017}.
The major departure from the standard MT setting
is that our model reformulates utterances in the same language.
Unlike in MT, there is little high
quality training data available for monolingual paraphrasing.
Effective training of highly parametrized neural networks relies on an
abundance of data.
We address this challenge by first pre-training on a related task, multilingual
translation, and then using signals produced during the interaction with the
environment for adaptation.

\subsection{Answer Selection Model}  \label{sec:ASModel}

During training, we have access to the reward for the answer returned
for each reformulation $q_i$.
However, at test time we must predict the best answer $a^*$.
The selection model selects the best answer from the set $\{a_i\}_{i=1}^N$
observed during the interaction by predicting the difference of the F1 score
to the average F1 of all variants.
We use pre-trained embeddings for the tokens
of query, rewrite, and answer. For each, we add a 1-dimensional CNN
followed by max-pooling. The three resulting vectors
are then concatenated and passed through a feed-forward network which
produces the output.
 \section{Training} \label{sec:training}

\subsection{Question Answering Environment}

We train a model on the training set for the QA task at hand, see
Section~\ref{sec:baselines} for details.
Afterwards, BiDAF becomes the black-box
environment and its parameters are not updated further.
In principle, we could train both the
agent and the environment jointly to further improve performance.
However, this is not our desired task: our aim is for the agent to
learn to communicate using natural language with an environment over
which is has no control.

\subsection{Policy Gradient Training of the Reformulation Model}

For a given question $q_0$, we want to return the best possible answer $a^*$,
maximizing a reward $a^*=\operatorname{argmax}_a R(a|q_0)$.
Typically, ${R}$ is the token level F1 score on the answer.
The answer $a = f(q)$ is an unknown function
of a question $q$, computed by the environment.
The reward is computed
with respect to the original question $q_0$ while the answer is provided for
$q$.
The question is generated according to a policy
$\pi_\theta$ where $\theta$ are the policy's parameters $q \sim \pi_\theta(\bigcdot|q_0)$.
The policy, in this case, a sequence-to-sequence model, assigns a probability
\begin{equation}
\pi_\theta(q|q_0) = \prod_{t=1}^Tp(w_t|w_1,\ldots,w_{t-1},q_0)
\end{equation}
to any possible question $q = w_1,\ldots,w_{T}$, where $T$ is the length of $q$
with tokens $w_t \in V$ from a fixed vocabulary $V$.
The goal is to maximize the expected reward of the answer returned under the policy,
$\mathbb{E}_{q\sim\pi_\theta(\bigcdot|q_0)}[{R}(f(q))]$. We
optimize the reward directly with respect to
parameters of the policy using Policy Gradient
methods~\citep{Sutton:1998:IRL:551283}.
The expected reward cannot be computed in closed form,
so we compute an unbiased estimate with Monte Carlo sampling,
\begin{equation}
\mathbb{E}_{q\sim\pi_\theta(\bigcdot|q_0)}[{R}(f(q))]
\approx \dfrac{1}{N} \sum_{i=1}^N {R}(f(q_i)),\quad q_i\sim\pi_\theta(\bigcdot|q_0)
\end{equation}

To compute gradients for training we use REINFORCE~\citep{Williams:1991},
\begin{align}
	\nabla\mathbb{E}_{q\sim\pi_\theta(\bigcdot|q_0)}[{R}(f(q))]
	&= \mathbb{E}_{q\sim\pi_\theta(\bigcdot|q_0)}\nabla_\theta
	\log(\pi_\theta(q|q_0))R(f(q))\\
	&\approx \dfrac{1}{N} \sum_{i=1}^N
	\nabla_\theta \log(\pi(q_i|q_0))R(f(q_i)),\quad q_i\sim\pi_\theta(\bigcdot|q_0)
\end{align}

This estimator is often found to have high variance, leading to
unstable training~\citep{greensmith2004variance}.
We reduce the variance by subtracting the following baseline reward:
$B(q_0)=\mathbb{E}_{q\sim\pi_\theta(\bigcdot|q_0)}[R(f(q))]$.
This expectation is also computed by sampling from the policy given $q_0$.

We often observed collapse onto a
sub-optimal deterministic policy. To address this we use entropy
regularization
\begin{equation}
  H[\pi_{\theta}(q|q_0)] = - \sum_{t=1}^T \sum_{w_t\in V} p_{\theta}(w_t|w_{<t},q_0)
  \log p_{\theta}(w_t|w_{<t},q_0) 
\end{equation}
This final objective is:
\begin{equation}
 \mathbb{E}_{q\sim\pi_\theta(\bigcdot|q_0)}[{R}(f(q)) - B(q_0)] + \lambda H[\pi(q|q_0)],
\end{equation}
where $\lambda$ is the regularization weight.

\subsection{Answer Selection}

Unlike the reformulation policy, we train the answer
with either beam search or sampling. We can produce
many rewrites of a single question from our reformulation system.
We issue each rewrite to the QA
environment, yielding a set of (query, rewrite, answer) tuples from which
we need to pick the best instance. We train another neural network to
pick the best answer from the candidates.
We frame the task as binary classification, distinguishing
between above and below average performance. In training, we
compute the F1 score of the answer for every instance.
If the rewrite produces an answer with an F1 score greater than the average
score of the other rewrites the instance is assigned a positive label.
We ignore questions where all rewrites yield equally good/bad answers.
We evaluated FFNNs, LSTMs, and CNNs and found that the
performance of all systems was comparable.
We choose a CNN which offers good
computational efficiency and accuracy (cf.\ \ref{sec:ASModel}).

\subsection{Pretraining of the Reformulation Model}

We pre-train the policy by building a paraphrasing Neural
MT model that can translate from English to English. While parallel
corpora are available for many language pairs, English-English corpora are
scarce.
We first produce a multilingual translation
system that translates between several languages~\citep{ZeroShot}. This
allows us to use available bilingual corpora. Multilingual training requires
nothing more than adding two special tokens to every line
which indicate the source and target languages.
The encoder-decoder architecture of the translation
model remains unchanged.

As~\citet{ZeroShot} show, this model can be used for \emph{zero-shot
translation}, i.e.\ to translate between language pairs for which it has seen no
training examples. For example, after training English-Spanish, English-French,
French-English, and Spanish-English the model has learned a single encoder that
encodes English, Spanish, and French and a decoder for the same three languages.
Thus, we can use the same model for French-Spanish, Spanish-French and also
English-English translation by adding the respective tokens to the source.~\citet{ZeroShot}
note that zero-shot translation usually performs worse than
bridging, an approach that uses the model twice: first, to translate
into a pivot language, then into the target language.
However, the performance gap can be
closed by running a few training steps for the desired language pair.
Thus, we first train on multilingual data, then on a small corpus of
monolingual data.

  \section{Experiments}

\subsection{Question Answering Data and BiDAF training}

SearchQA~\citep{Dunn:2017} is a dataset built
starting from a set of
\emph{Jeopardy!}\ clues. Clues are obfuscated queries such as
\emph{This `Father of Our Country' didn't really chop down a cherry tree}. Each clue
is associated with the correct answer, e.g.\ \emph{George Washington},
and a list of snippets from Google's top search
results. SearchQA contains over 140k question/answer pairs and 6.9M snippets.
We train our model on the pre-defined training split,
perform model selection and tuning on the
validation split and report results on the validation and test splits.
The training, validation and test sets contain 99,820,
13,393 and 27,248 examples, respectively.

We train BiDAF directly on the SearchQA training data.
We join snippets to form the
context from which BiDAF selects answer spans.
For performance reasons, we limit the context to the top 10 snippets.
This corresponds to finding the answer on the first page of Google results.
The results are only mildly affected by this limitation,
for 10\% of the questions, there is no answer in this shorter context.
These data points are all counted as losses.
We trained with the Adam optimizer for 4500 steps, using learning rate 0.001,
batch size 60.

\subsection{Question Reformulator Training}
\label{nmt-pretraining}
For the pre-training of the reformulator, we use the
multilingual United Nations Parallel Corpus v1.0~\citep{unv1}.
This dataset contains 11.4M sentences which are fully
aligned across six UN languages:
Arabic, English, Spanish, French, Russian, and Chinese.
From all bilingual pairs, we produce a multilingual training
corpus of 30 language pairs. This yields 340M training examples
which we use to train the zero-shot neural MT system~\citep{ZeroShot}.
We tokenize our data using 16k sentence pieces.\footnote{
\url{https://github.com/google/sentencepiece}}
Following \citet{Britz:2017} we use a
bidirectional LSTM as the encoder and a 4-layer
stacked LSTM with attention as the decoder.
The model converged after training on 400M instances using the Adam optimizer
with a learning rate of 0.001 and batch size of 128.

The model trained as described above has poor
quality. For example,
for the question \emph{What month, day and year did Super Bowl 50 take
place?}, the top
rewrite is \emph{What month and year goes back to the morning and year?}.
To improve quality, we resume training on a smaller monolingual dataset,
extracted from the Paralex database of question
paraphrases~\citep{Fader:2013}.\footnote{
\url{http://knowitall.cs.washington.edu/paralex/}}
Unfortunately, this data contains many noisy pairs.
We filter many of these pairs out by keeping only those where the Jaccard
coefficient between the sets of source and target terms is above 0.5.
Further, since the number of paraphrases for each
question can vary significantly, we keep at most 4
paraphrases for each question.
After processing, we are left with about 1.5M pairs out of the original
35M.
The refined model has visibly better quality
than the zero-shot one; for the example question above it
generates \emph{What year did superbowl take place?}.
We also tried training
on the monolingual pairs alone. As in~\citep{ZeroShot}, the
quality was
in between the multilingual and refined models.

After pre-training the reformulator, we switch the optimizer from Adam to SGD
and train for $100\text{k}$ RL steps of batch size $64$ with a low learning
rate of $0.001$. We use an entropy regularization weight of $\lambda=0.001$.
For a stopping criterion, we monitor the reward from the best single rewrite,
generated via greedy decoding, on the validation set.
In contrast to our initial training which we ran on GPUs, this training phase is
dominated by the latency of the QA system and we run inference and updates on
CPU and the BiDAF environment on GPU.

\subsection{Training the Answer Selector}

For the selection model we use supervised learning: first, we train the reformulator,
then we generate $N=20$ rewrites for each question in
the SearchQA training and validation sets.
After sending these to the environment we
have about 2M (question, rewrite, answer) triples.
We remove queries where all rewrites yield identical rewards, which removes
about half of the training data.
We use pre-trained 100-dimensional embeddings~\citep{glove} for the tokens.
Our CNN-based selection model encodes the three strings into 100-dimensional
vectors using a 1D CNN with kernel width 3 and output dimension 100 over the
embedded tokens, followed by max-pooling. The vectors are then concatenated
and passed through a feed-forward network which produces the binary output,
indicating whether the triple performs below or above average, relative to the
other reformulations and respective answers.

We use the training portion of the SearchQA data thrice, first
for the initial training of the BiDAF model, then for the reinforcement-learning
based tuning of the reformulator, and finally for the training of the selector.
We carefully monitored
that this didn’t cause severe overfitting. BiDAF alone has a
generalization gap between the training and validation set errors of
3.4 F1. This gap remains virtually identical after training the
rewriter. After training the CNN, AQA-Full has a slightly larger gap
of 3.9 F1. We conclude that training AQA on BiDAF’s training set
causes very little additional overfitting.
We use the test set only for
evaluation of the final model.

\subsection{Baselines and Benchmarks} \label{sec:baselines}
As a baseline, we report the results of the modified pointer
network, called Attention Sum Reader (ASR), developed for SearchQA
\citep{Dunn:2017}.
We also report the performance of the BiDAF environment used without
the reformulator to answer the original question.

We evaluate against several benchmarks. First,
following~\citet{Kumaran:2009}, we implement a system (MI-SubQuery)
that generates
reformulation candidates by enumerating all subqueries of the original
SearchQA query and then keeps the top $N$ ranked by mutual
information.\footnote{As in~\citep{Kumaran:2009}, we choose the reformulations
  from the term-level subsequences of length 3 to 6. We associate each
  reformulation with a graph, where the vertices are the terms and the edges are
  the mutual information between terms with respect to the collection of
  contexts. We rank the reformulations by the average edge weights of Maximum
  Spanning Trees of the corresponding graphs.}
From this set, we pick the highest scoring one as the
top hypothesis to be used as a single rewrite. We also use the whole
set to train a CNN answer selector for this specific source of
rewrites. In this way, we can compare systems fairly both in single
prediction or ensemble prediction modes. Additionally, we evaluate
against another source of reformulations: the zero-shot monolingual NMT
system trained on the U.N. corpus and Paralex (Base-NMT), without reinforcement
learning. As with the MI-SubQuery benchmark, we evaluate the
Base-NMT system both as a single reformulation predictor and as a
source of $N$ best rewrites, for which we train a dedicated CNN answer
selector. We also report human performance on SearchQA, based on a sample of the
test set, from~\citep{Dunn:2017}.

\subsection{Results}
\begin{table}[t]
  \centering
    \begin{tabular}{lllllllllllll}
    \toprule
    &&\multicolumn{2}{c}{Baseline}& \multicolumn{2}{c}{MI-SubQuery} &\multicolumn{2}{c}{Base-NMT} &\multicolumn{4}{c}{AQA} \\
     \cmidrule(lr){3-4} \cmidrule(lr){5-6} \cmidrule(lr){7-8} \cmidrule(lr){9-12}
    && \rot{ASR} & \rot{BiDAF} & \rot{TopHyp} & \rot{CNN} & \rot{TopHyp} & \rot{CNN} & \rot{TopHyp} & \rot{Voting} & \rot{MaxConf} & \rot{CNN} & \rot{Human}\\\midrule
    \multirow{2}[3]{*}{Dev}  & EM & -    & 31.7 & 24.1 & 37.5 & 26.0 &
    37.5 & 32.0 & 33.6 & 35.5 & {\bf 40.5} & -\\
         & F1 & 24.2 & 37.9 & 29.9 & 44.5 & 32.2 & 44.8 & 38.2 & 40.5
    & 42.0 & {\bf 47.4} & -\\
    \multirow{2}[3]{*}{Test} & EM & -    & 28.6 & 23.2 & 35.8 & 24.8 & 35.7 & 30.6 & 33.3 & 33.8 & {\bf 38.7} & 43.9\\
    & F1 & 22.8 & 34.6 & 29.0 & 42.8 & 31.0 & 42.9 & 36.8 & 39.3 &
    40.2 & {\bf 45.6} & -\\
    \bottomrule
  \end{tabular}
  \caption{Results table for the experiments on SearchQA.
   Two-sample \emph{t}-tests between the AQA results and either the
   Base-NMT or the MI-SubQuery results show that differences in F1 and
   Exact Match scores are statistically significant, $p<10^{-4}$, for
   both Top Hypothesis and CNN predictions. The difference between
   Base-NMT and MI-SubQuery is also significant for Top Hypothesis
   predictions.} 
  \label{tab:results}
\end{table}

We evaluate several variants of AQA. For each query $q$ in
the evaluation we generate a list of reformulations $q_{i}$,
for $i=1\ldots N$, from the
AQA reformulator trained as described in Section~\ref{sec:training}.
We set $N=20$ in these experiments, the same value is used for the benchmarks.
In \emph{AQA TopHyp} we use
the top hypothesis generated by the sequence model, $q_1$.
In \emph{AQA Voting} we use BiDAF scores for a heuristic weighted
voting scheme to implement deterministic selection.
Let $a$ be the answer returned
by BiDAF for query $q$, with an associated score $s(a)$.
We pick the answer according to
$\operatorname{argmax}_{a} \sum_{a^{\prime}=a} s(a^{\prime})$.
In \emph{AQA MaxConf} we select
the answer with the single highest BiDAF score across question reformulations.
Finally, \emph{AQA CNN} identifies the complete system with the learned
CNN model described in Section~\ref{sec:AQAmodel}.

Table~\ref{tab:results} shows the results. We report exact match (EM)
and F1 metrics, 
computed on token level between the predicted answer and the gold answer.
We present results on the full validation and test
sets (referred to as  \emph{$n$-gram} in \citep{Dunn:2017}).
Overall, SearchQA appears to be harder than other recent
QA tasks such as SQuAD~\citep{Rajpurkar:2016}, for both machines and humans.
BiDAF's performance drops by 40 F1
points on SearchQA compared to SQuAD.
However, BiDAF is still competitive on SeachQA, improving over
the Attention Sum Reader network by 13.7 F1 points.

Using the top hypothesis already yields an improvement of 2.2 F1 on the
test set.
This demonstrates that even the reformulator alone
is capable to produce questions more easily answered by the
environment. When generating a single prediction, both MI-SubQuery and
Base-NMT benchmarks perform worse than BiDAF.
Heuristic selection via both Voting and Max Conf yields a further performance
boost. Both heuristics draw upon the intuition that when BiDAF is
confident in its answer it is more likely 
to be correct, and that multiple instances of the same answer provide positive
evidence (for MaxConf, the max operation implicitly rewards having an answer
scored with respect to multiple questions).
Finally, a
trained selection function improves performance further, yielding an absolute
increase of 11.4 F1 points (32\% relative) over BiDAF with the original
questions. In terms of exact match score, this more than closes half the gap
between BiDAF 
and human performance. The benchmarks improve considerably when they
generate $N$ candidates, and paired with a dedicated CNN
selector. This is not surprising as it provides an ensemble prediction
setup. However, the AQA CNN system outperforms both MI-SubQuery and
Base-NMT in all conditions by about 3\%.

Finally, we consider the maximum performance possible
that could be achieved by picking the answer with the highest F1 score
from the set of those returned for all available reformulations.
Here we find that the different sources of
rewrites provide comparable headroom: the oracle Exact Match is near
50, while the oracle F1 is close to 58.

 \section{Analysis of the agent's language}
The AQA agent can learn several types of sub-optimal policies. For example,
it can converge to deterministic policies by learning to
emit the same, meaningless, reformulation for any input question. This
strategy can lead to local optima because the environment has
built in strong priors on what looks like a likely answer, even
ignoring the input question. Hence, convergence to non-negligible
performance is easy. Entropy regularization typically
fixes this behavior. Too much weight on the entropy regularizer, on the
other hand, might yield random policies. A more
competitive sub-optimal policy is one that generates  
minimal changes to the input, in order to stay close to the
original question. This is a successful strategy because the
environment has been trained on the original questions alone, which
leads to baseline performance.

It seems quite remarkable then that AQA is able to learn non-trivial
reformulation policies, that differ significantly from all of the
above. One can think of the policy as a language for 
formulating questions that the agent has developed while engaging in a
machine-machine communication with the environment. In this section we
look deeper into the agent's language.

\subsection{General properties}
We analyze input questions
and reformulations on the  
development partition of SearchQA to gain insights on how the agent's language
evolves during training via policy gradient. It is important to note that in the
SearchQA dataset the original \emph{Jeopardy!} clues have been preprocessed by
lower-casing and stop word removal. The resulting preprocessed clues that form
the sources (inputs) for the sequence-to-sequence reformulation model resemble
more keyword-based search queries than grammatical questions. For example, the
clue \emph{Gandhi was deeply influenced by this count who wrote "War and Peace"}
is simplified to \emph{gandhi deeply influenced count wrote war peace}.

The (preprocessed) SearchQA questions contain 9.6 words on average.
They contain few repeated terms, computed as the mean term frequency (TF) per
question.
The average is 1.03, but for most of the queries (75\%) TF is
1.0. We also compute the median
document frequency (DF) per query, where the document is the
context from which the answer is
selected, as a measure of how informative a term is.\footnote{For
  every token we compute the number of contexts
  containing that token. Out of these counts, in
  each question, we take the median instead of the mean to reduce the
  influence of frequent outliers, such as commas.} As another measure of
query performance, we also compute Query Clarity
(QC)~\citep{Cronen-Townsend:2002}.\footnote{The relative
entropy between a query language model and the corresponding
collection language model. In our case, the document is the context for each
query.}
\begin{figure}
\begin{center}
\includegraphics[trim={1cm 1cm 0 1cm},clip,width=0.9\linewidth]{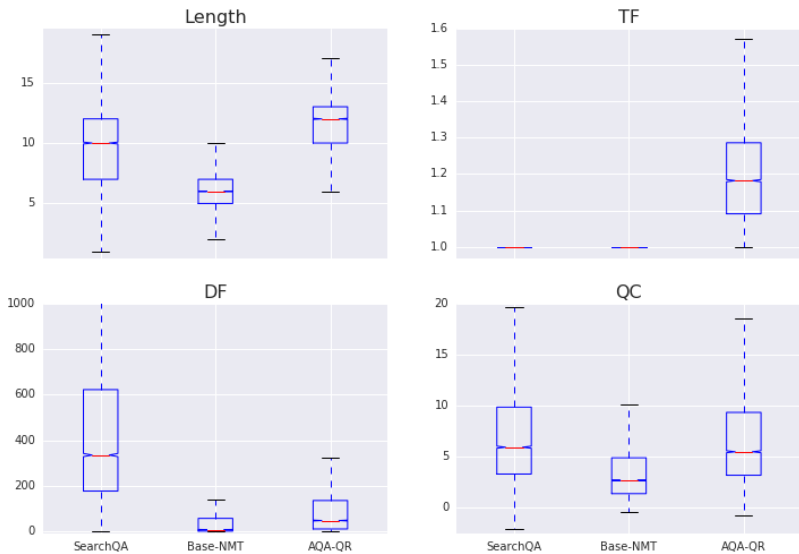}
\caption{Boxplot summaries of the statistics collected for all types of
  questions. Two-sample \emph{t}-tests performed on all pairs in
  each box confirm that the differences in means are significant $p<10^{-3}$.}
\label{fig:boxplot}
\end{center}
\end{figure}
Figure~\ref{fig:boxplot} summarizes statistics of the
questions and rewrites.

We first consider the top hypothesis generated by the pre-trained
NMT reformulation system, before reinforcement learning (Base-NMT).
The Base-NMT rewrites differ greatly from their sources.
They are shorter, 6.3 words
on average, and have even fewer repeated terms (1.01).
Interestingly, these reformulations are mostly syntactically well-formed
questions. For example, the
clue above becomes \emph{Who influenced count wrote war?}.\footnote{More
examples can be found in Appendix~\ref{app1}.}
Base-NMT improves structural language quality by properly reinserting
dropped function words and wh-phrases.
We also verified the increased
fluency by using a large language model and found that the Base-NMT
rewrites are 50\% more likely than the original questions.
While more fluent, the Base-NMT rewrites involve lower DF terms.
This is probably due to a domain mismatch
between SearchQA and the NMT training corpus. The query clarity of the
Base-NMT rewrites is also degraded as a result of the transduction process.

We next consider the top hypothesis generated by the AQA question
reformulator (AQA-QR) after the policy gradient training.
The AQA-QR rewrites are those whose
corresponding answers are evaluated as
\emph{AQA TopHyp} in Table~\ref{tab:results}. These single
rewrites alone 
outperform the original SearchQA queries by 2\% on the test set.
We analyze the top hypothesis instead of the final output of the full
AQA agent to avoid confounding effects from the answer selection step.
These rewrites look
different from both the Base-NMT and the SearchQA
ones. For the example above AQA-QR's top hypothesis
is \emph{What is name gandhi gandhi influence wrote peace
peace?}. Surprisingly, 99.8\% start with the
prefix \emph{What is name}. The second most frequent is
\emph{What country is} (81 times), followed by \emph{What is is} (70) and
\emph{What state} (14). This is puzzling as it occurs in only 9
Base-NMT rewrites, and never in the original SearchQA questions.
We speculate it might be related to the fact that
virtually all answers involve names,
of named entities (Micronesia) or generic concepts (pizza).

AQA-QR's rewrites seem less fluent than both the
SearchQA and the Base-MT counterparts.
In terms of language model probability, they are less likely than both SearchQA
and Base-NMT.
However, they have more repeated terms (1.2 average TF),
are significantly longer (11.9) than in Base-NMT and contain more
informative context terms than SearchQA questions (lower DF). Also, the
translation process does not affect query clarity much.
Finally, we find that AQA-QR's reformulations contain
morphological variants in 12.5\% of cases.
The number of questions that
contain multiple tokens with the same stem doubles from SearchQA to
AQA-QR. Singular forms are preferred over plurals. Morphological
simplification is useful because it increases the chance
that a word variant in the question matches the context.

\subsection{Paraphrasing quality}
We also investigate the general paraphrasing abilities of our model, focusing on
the relation between paraphrasing quality and QA quality. To tease apart the
relationship between paraphrasing and reformulation for QA we evaluated 3
variants of the reformulator:

{\bf Base-NMT}\quad  This is the model used to initialize RL training of the
agent. Trained first on the multilingual U.N. corpus, then on the Paralex
corpus, as detailed in Section \ref{nmt-pretraining}.\\
{\bf Base-NMT-NoParalex}\quad  This is the model above trained
solely on the multilingual U.N. corpus, without the Paralex monolingual
corpus.\\
{\bf Base-NMT+Quora}\quad This is the same as
Base-NMT, additionally trained on the Quora
dataset\footnote{\url{https://data.quora.com/First-Quora-Dataset-Release-Question-Pairs}}
which contains 150k duplicate questions.\\

Following \citet{Prakash:2016}, we evaluate all models on the
MSCOCO\footnote{\url{http://cocodataset.org/}} \citep{MSCOCO:2014} validation set
(VAL2014). This dataset consists of images with 5 captions each, of which we
select a random one as the source and the other four as references. We use beam
search, to compute the top hypothesis and report uncased,
moses-tokenized BLEU using multeval\footnote{\url{https://github.com/jhclark/multeval}} \citep{Clark:2011}.
Please note, that the MSCOCO data is only used for evaluation
purposes. Examples of all systems can be found in Appendix \ref{mscocoexamples}.

The Base-NMT model performs at 11.4 BLEU (see Table \ref{tab:results} for the QA
eval numbers). In contrast, Base-NMT-NoParalex performs poorly at 5.0 BLEU.
Limiting training to the multilingual data alone also degrades QA performance:
the scores of the Top Hypothesis are at least 5 points lower in all metrics and
CNN scores are 2-3 points lower.

By training on additional monolingual data, the
Base-NMT+Quora model improves BLEU score slightly to 11.6. End-to-end QA
performance also improves marginally, the maximum delta with respect to Base-NMT
under all conditions is +0.5 points, but the difference is not statistically
significant. Thus, adding the Quora training does not have a significant effect.
This might be due to the fact that most of the improvement is captured by
training on the larger Paralex data set. Improving raw paraphrasing quality as
well as reformulation fluency helps AQA up to a point. However, they are only
partially aligned with the main task, which is QA performance. The AQA-QR
reformulator has a BLEU score of 8.6, well below both Base-NMT models trained on
monolingual data. Yet, AQA-QR significantly outperforms all others in the QA
task. Training the agent starting from the Base-NMT+Quora model yielded
comparable results as starting from Base-NMT.

\subsection{Discussion}
Recently, \citet{Lewis:2017} trained chatbots
that negotiate via language utterances in order to complete a
task. They report that the
agent's language diverges from human language if there is no
incentive for fluency in the reward function.
Our findings seem related.
The fact that the questions reformulated by AQA do not
resemble natural language is not due to the keyword-like
SearchQA input questions, because Base-NMT is capable of producing
more fluent
questions from the same input.
AQA learns
to re-weight terms by focusing on informative (lower document frequency),
query-specific (high query clarity), terms
while increasing term frequency (TF) via duplication. At the same time
it learns to modify surface forms in ways akin to stemming and
morphological analysis.

Some of the techniques seem to
adapt to the specific properties of current deep QA architectures
such as character-based modeling and attention.
Sometimes AQA learns to generate
semantically nonsensical, novel, surface term variants; e.g., it
might transform the adjective \emph{dense} to \emph{densey}. The only
justification for this is that such forms can be still exploited by the
character-based BiDAF question encoder.
Finally, repetitions can directly increase
the chances of alignment in the attention components.

We hypothesize that, while there is no
incentive for the model to use human language
due to the nature of the task,
AQA learns to ask BiDAF questions by
optimizing a language that increases the likelihood of BiDAF
\emph{ranking} better the candidate answers.
\citet{Jia:2017} argue that reading comprehension
systems are not capable of significant language understanding and
fail easily in adversarial settings.
We speculate that current machine comprehension tasks involve mostly
pattern matching and relevance modeling. As a consequence
deep QA systems might implement sophisticated
ranking systems trained to sort snippets of text from the context.
As such, they resemble document retrieval systems
which incentivizes the (re-)discovery
of IR techniques, such as tf-idf re-weighting and stemming, that have
been successful for decades~\citep{Baeza-Yates:1999}.

\section{Conclusion}
We propose a new framework to improve question answering.
We call it active question answering (AQA), as it aims
to improve answering by systematically perturbing input questions.
We investigated a first system of this kind that has three
components: a question reformulator, a black box QA system, and a
candidate answer aggregator.
The reformulator and aggregator form a trainable agent that seeks to elicit the
best answers from the QA system. Importantly, the agent may only query
the environment with natural language questions.
Experimental results prove that the approach is highly effective and
that the agent is able to learn non-trivial and somewhat interpretable
reformulation policies.

For future work, we will continue developing active question
answering, investigating the
sequential, iterative aspects of information seeking tasks, framed
as end-to-end RL problems, thus, closing the loop
between the reformulator and the selector.

 \section{Acknowledgements}
We would like to thank the anonymous reviewers for their valuable
comments and suggestions. We would also like to thank Jyrki
Alakuijala, G\'{a}bor B\'{a}rtok, Alexey Gronskiy,
Rodrigo Nogueira and Hugo Penedones
for insightful discussions and technical feedback.
 
\newpage
\begingroup
\bibliography{aqa}
\bibliographystyle{iclr2018_conference}
\endgroup

\newpage
\appendix
\section{Reformulation Examples}\label{appa}\label{app1}
  \scriptsize
  \begin{longtable}{r|p{6.5cm}|p{5.1cm}}
  \caption{Results of the qualitative analysis on SearchQA. For the original \emph{Jeopardy!} questions we give the reference answer, otherwise the answer given by BiDAF.} \\
\toprule
    {\bf Model} & {\bf Query} & {\bf Reference / Answer from BiDAF (F1)} \\
    \midrule
    Jeopardy! & People of this nation AKA Nippon wrote with a brush, so painting became the preferred form of artistic expression & japan\\
    SearchQA  & people nation aka nippon wrote brush , painting became preferred form artistic expression & japan (1.0)\\
    MI        & nippon brush preferred & julian (0)\\
    Base-NMT  & Aka nippon written form artistic expression? & julian (0)\\
    AQA-QR    & What is name did people nation aka nippon wrote brush expression? & japan (1.0)\\
    AQA-Full  & people nation aka nippon wrote brush , painting became preferred form artistic expression & japan (1.0)\\
    \midrule
    Jeopardy! & Michael Caine \& Steve Martin teamed up as Lawrence \& Freddy, a couple of these, the title of a 1988 film & dirty rotten scoundrels\\
    SearchQA  & michael caine steve martin teamed lawrence freddy , couple , title 1988 film & dirty rotten scoundrels (1.0)\\
    MI        & caine teamed freddy & dirty rotten scoundrels (1.0) \\
    Base-NMT  & Who was lawrence of michael caine steve martin? & rain man 1988 best picture fikkle [...~25~tokens] (0.18) \\
    AQA-QR    & What is name is name is name michael caine steve martin teamed lawrence freddy and title 1988 film? & dirty rotten scoundrels (1.0)\\
    AQA-Full  & What is name is name where name is name michael caine steve martin teamed lawrence freddy and title 1988 film key 2000 ? & dirty rotten scoundrels (1.0) \\
    \midrule
    Jeopardy!  & Used underwater, ammonia gelatin is a waterproof type of this explosive & dynamite \\
    SearchQA   & used underwater , ammonia gelatin waterproof type explosive & nitroglycerin (0)\\
    MI         & ammonia gelatin waterproof & nitroglycerin (0)\\
    Base-NMT   & Where is ammonia gelatin waterproof? nitroglycerin (0)\\
    AQA-QR     & What is name is used under water with ammonia gelatin water waterproof type explosive? & nitroglycerin (0)\\
    AQA-Full   & used underwater , ammonia gelatin waterproof type explosive & nitroglycerin (0)\\
    \midrule
    Jeopardy! & The Cleveland Peninsula is about 40 miles northwest of Ketchikan in this state & alaska\\
    SearchQA  & cleveland peninsula 40 miles northwest ketchikan state & alaska 's community information summary says [... 113 tokens] (0.02)\\
    MI        & cleveland peninsula ketchikan & alaska 's dec 16 , 1997 [...~132~tokens] (0.01)\\
    Base-NMT  & The cleveland peninsula 40 miles? & ketchikan , alaska located northwest tip \mbox{[...~46~tokens]} (0.04)\\
    AQA-QR    & What is name is cleveland peninsula state northwest state state state? & alaska (1.0)\\
    AQA-Full  & What is name are cleveland peninsula state northwest state state state ? & alaska (1.0)\\
    \midrule
    Jeopardy! & Tess Ocean, Tinker Bell, Charlotte the Spider & julia roberts\\
    SearchQA  & tess ocean , tinker bell , charlotte spider & julia roberts tv com charlotte spider [...~87~tokens] (0.04) \\
    MI        & tess tinker spider & julia roberts tv com charlotte spider [...~119~tokens] (0.01)\\
    Base-NMT  & What ocean tess tinker bell? & julia roberts american actress producer made [...~206~tokens] (0.02)\\
    AQA-QR    & What is name tess ocean tinker bell link charlotte spider? & julia roberts (1.0)\\
    AQA-Full  & What is name is name tess ocean tinker bell spider contain charlotte spider contain hump around the world winter au to finish au de mon moist & julia roberts (1.0)\\
    \midrule
    Jeopardy! & During the Tertiary Period, India plowed into Eurasia \& this highest mountain range was formed & himalayas\\
    SearchQA  & tertiary period , india plowed eurasia highest mountain range formed & himalayas (1.0)\\
    MI        & tertiary plowed eurasia & himalayas (1.0) \\
    Base-NMT  & What is eurasia highest mountain range?  & himalayas (1.0)\\
    AQA-QR    & What is name were tertiary period in india plowed eurasia? & himalayas (1.0)\\
    AQA-Full  & tertiary period , india plowed eurasia highest mountain range formed & himalayas (1.0)\\
    \midrule
    Jeopardy!  & The melody heard here is from the opera about Serse, better known to us as this "X"-rated Persian king & xerxes\\
    SearchQA   & melody heard opera serse , better known us x rated persian king & gilbert sullivan (0)\\
    MI         & melody opera persian & gilbert sullivan (0)\\
    Base-NMT   & Melody heard opera serse thing? & gilbert sullivan (0)\\
    AQA-QR     & What is name melody heard opera serse is better persian king? & gilbert sullivan (0)\\
    AQA-Full   & What is name is name melody heard opera serse is better persian king persian K ? & gilbert sullivan (0)\\
   \bottomrule
  \end{longtable}
\pagebreak
\section{Examples of Ranking Losses}\label{oracle_examples}
  \begin{table*}[h!]
  \footnotesize
  \caption{Examples of queries where none of the methods produce the right answer, but the Oracle model can.} \vspace{3mm}
  \begin{tabular}{r|p{8cm}|r}
    \toprule
    {\bf Model} & {\bf Query} & {\bf Answer (F1)} \\
    \midrule
    SearchQA & ancient times , pentelic quarries major source building material athens & parthenon (0.0) \\
    AQA-QR & What is name an Ancient History Warry quarry material athens? & parthenon (0.0) \\
    AQA-CNN & ancient times , pentelic quarries major source building material athens & parthenon (0.0) \\
    AQA-Oracle & What is name is name an Ancient Romes For pentelic quarry material athens measure athens? & marble (1.0) \\
    \midrule
    SearchQA & 1949 1999 germany 's bundestag legislature met city & berlin (0.0) \\
    AQA-QR & What is name is name 1999 germany\'s bundestag legislature met city city? & berlin (0.0) \\
    AQA-CNN & What is name is name 1999 germany germany bundestag legislature city city? & berlin (0.0) \\
    AQA-Oracle & What is name is name a 1999 germany germany legislature met city city? & bonn (1.0) \\
    \midrule
    SearchQA & utah jazz retired 12 jersey & karl malone (0.0) \\
    AQA-QR & What is name did utah jazz retired jersey jersey? & karl malone (0.0) \\
    AQA-CNN & What is name did utah jazz retired jersey jersey? & karl malone (0.0) \\
    AQA-Oracle & What is name is name where utah jazz is retired jersey jersey? & john stockton (1.0) \\
    \midrule
    SearchQA & swing intersection count basie capital virginia & chicago (0.0) \\
    AQA-QR & What is name for swing intersection count basie capital virginia? & chicago (0.0) \\
    AQA-CNN & swing intersection count basie capital virginia & chicago (0.0) \\
    AQA-Oracle & What is name swing swing intersection capital virginia? & richmond (1.0) \\
    \midrule
    SearchQA & huge upset virginia , jim webb beat senator prospective presidential candidate & hillary clinton (0.0) \\
    AQA-QR & What is name is name huge upset virginia jug webb beat senator? & jim webb (0.0) \\
    AQA-CNN & What is name is name huge upset virginia jug webb beat senator world candidate? & jim webb (0.0) \\
    AQA-Oracle & What is name is name huge upset virginia  jim webb beat senator? & george allen (1.0) \\
    \bottomrule
\end{tabular}
\end{table*}
\vfill

\pagebreak
\section{Paraphrasing Examples}\label{mscocoexamples}
\begin{table*}[h!]
\footnotesize
  \caption{Paraphrasing examples on captions from MSCOCO \citep{MSCOCO:2014}.} \vspace{3mm}
  \centering
  \ra{1.0}
  \begin{tabular}{r|p{10cm}}
    \toprule
Source & a man riding a wave on top of a surfboard.\\
Reference & a man on a surfboard in the crest of a wave.\\
Base-NMT & a man on top of a surfboard?\\
Base-NMT-NoParalex & a mankind arises away from a floor.\\
Base-NMT+Quora & it possible to riding a wave on a surfboard?\\
AQA-QR & what is name an average man riding a wave at onfboard?\\
\midrule
Source & a person riding some skis through a snowy mountain trail.\\
Reference & a man on skis is standing by some trees.\\
Base-NMT & what is some skis through a snowy mountain trail?\\
Base-NMT-NoParalex & the person rising somekises through a snow blood.\\
Base-NMT+Quora & what is a skis mountain trail?\\
AQA-QR & what is name an average person riding some one snowy mountain trail?\\
\midrule
Source & a group of people all taking a bite out of the same sandwich.\\
Reference & a group of people eating rom a large sandwich.\\
Base-NMT & what do people call a bite out of the same sandwich?\\
Base-NMT-NoParalex & a group of people who takes a bargained from the same index.\\
Base-NMT+Quora & what is the group of people who take bite out of the same sandwich?\\
AQA-QR & what is name a group of people with bite out of sandwich?\\
\midrule
Source & a partially lit asian person holding a phone  \\
Reference & a man standing next to a sign talking on a phone.\\
Base-NMT & what is a partical that holds a phone?\\
Base-NMT-NoParalex & plus little as an associate person\\
Base-NMT+Quora & what is a partially little asian person?\\
AQA-QR & what is name is associated with a partially active asian person holding a phone?\\
\midrule
Source & a car driving through a very flooded street\\
Reference & a car is driving through a flooded street.\\
Base-NMT & what car driving through a flooded street?\\
Base-NMT-NoParalex & analysing through a heat street\\
Base-NMT+Quora & i driving through a flooded street?\\
AQA-QR & what is name is given to a car with a car driving through a flooded street?\\
\midrule
Source & a man is holding a tennis racket on a court\\
Reference & a man holding a racquet on top of a tennis court.\\
Base-NMT & how do you hold a tennis racket?\\
Base-NMT-NoParalex & a man to hold a threshold on a court\\
Base-NMT+Quora & what is a tennis racket?\\
AQA-QR & what is name an average man make a tennis racket on a court?\\
\midrule
Source & a pizza on a pan that is half gone\\
Reference & a table with a half eaten pizza and glasses if beer.\\
Base-NMT & what is half gone in a pizza?\\
Base-NMT-NoParalex & average on a size\\
Base-NMT+Quora & is a pizza on a pan that is half gone?\\
AQA-QR & what is name a pizza on a pizza pan that is half gone?\\
\midrule
Source & a boy holding a glove while on a baseball field.\\
Reference & a young boy bending over holding a catchers mitt.\\
Base-NMT & how do you hold a glove on a baseball field?\\
Base-NMT-NoParalex & a cell holding a globe at a later field.\\
Base-NMT+Quora & is it possible to hold a glove while on a baseball field?\\
AQA-QR & what is name an boy make up a glove while on baseball field?\\
\midrule
Source & a street light with a red arrow pointing left.\\
Reference & a tagged traffic light displays a red arrow.\\
Base-NMT & a street light with red arrow pointing?\\
Base-NMT-NoParalex & a strength with a skilling point.\\
Base-NMT+Quora & what is the street light of a red arrow pointing?\\
AQA-QR & what is name a street light with red arrow pointing left?\\
   \bottomrule
  \end{tabular}
\end{table*}

 \end{document}